\documentclass[letterpaper]{article}

\usepackage[preprint]{aaai2027}
\usepackage[hyphens]{url}
\usepackage{graphicx}
\usepackage{natbib}
\usepackage{caption}
\usepackage{booktabs}
\usepackage{multirow}
\usepackage{amsmath}
\usepackage{amssymb}
\usepackage{xspace}
\usepackage{pifont}

\title{HyperClaim: Fine-Grained Cross-Modal Hypergraph Reasoning for Video Misinformation Detection}
\author{
    Xiangbo Wang\textsuperscript{\rm 1},
    Jiasheng Zhang\textsuperscript{\rm 2},
    Xingtong Yu\textsuperscript{\rm 3},
    Luoqiang Lei\textsuperscript{\rm 1},
    Delvin Ce Zhang\textsuperscript{\rm 4}\corresponding
}
\affiliations{
    \textsuperscript{\rm 1}School of Communication Engineering, Hangzhou Dianzi University\\
    \textsuperscript{\rm 2}School of Computer Science and Technology, Xidian University\\
    \textsuperscript{\rm 3}School of Computing and Information Systems, Singapore Management University\\
    \textsuperscript{\rm 4}School of Computer Science, University of Sheffield; delvin.ce.zhang@sheffield.ac.uk
}

\newcommand{\method}{HyperClaim\xspace}
\newcommand{\hforge}{\textsc{H-Forge}\xspace}
\newcommand{\aether}{\textsc{Aether}\xspace}
\newcommand{\cred}{\textsc{Cred}\xspace}
\newcommand{\realnews}{\textsc{Real}\xspace}
\newcommand{\fakenews}{\textsc{Fake}\xspace}
\newcommand{\cmark}{\ding{51}}
\newcommand{\xmark}{\ding{55}}
\newcommand{\ablmetric}[1]{\multicolumn{1}{c}{#1}}
\newcommand{\defaultmark}{\textsuperscript{\ensuremath{\dagger}}}

\begin{document}

\maketitle

\begin{abstract}
Video misinformation detection is often approached through global multimodal fusion or free-form multimodal reasoning. Both paradigms can under-represent localized authenticity cues that arise from coupled interactions among query phrases, contextual text, and short temporal spans of frames. Because such interactions are inherently higher-order, pairwise graph formulations are insufficient to capture multi-way cross-modal dependencies, whereas hypergraphs offer a suitable representation for these relations. We propose HyperClaim, a discriminative temporal hypergraph framework for sample-level authenticity classification. Using the title or benchmark-provided paired text as a claim-like query, HyperClaim constructs a sparse heterogeneous hypergraph over query tokens, evidence tokens, and sampled frames; applies confidence-aware filtering and source budgeting to form compact text--frame and short-range temporal evidence units; performs adaptive soft-incidence reasoning with residual text--video calibration; and aggregates textual, visual, and hyperedge states through a discrepancy-aware readout. Without relying on generated rationales or external tool calls, HyperClaim preserves fine-grained cross-modal and temporal structure that global fusion tends to flatten. Under the FactGuard temporal protocol, it achieves 83.7\%, 82.0\%, and 87.3\% accuracy on FakeSV, FakeTT, and FakeVV, respectively, outperforming strong discriminative and reasoning-centric baselines. Learned incidence and attention weights further reveal token- and frame-level structural evidence paths.
\end{abstract}

\section{Introduction}

Short-form video has become a major channel for online news and public discourse~\cite{qi2023fakesv,bu2024fakingrecipe}. Its combination of language, imagery, audio-derived text, and temporal editing also enables misinformation that is difficult to identify from any single modality. Misleading content may reuse authentic footage in a new context, modify an entity or location in the accompanying text, or present a short segment under an unsupported narrative~\cite{luo2021newsclippings,papadopoulos2024verite,bu2024fakingrecipe,zhang2025factr1}. In such cases, most of the sample may remain plausible while the decisive evidence is confined to a title phrase, a small set of textual cues or frames, or a brief temporal transition. Other cases rely on broader social, contextual, or editing signals~\cite{qi2023fakesv,qi2023need,bu2024fakingrecipe}. Effective detection should therefore preserve localized cross-modal evidence without assuming that every misleading sample reduces to a single explicit contradiction.

\begin{figure}[t]
    \centering

    \includegraphics[width=0.78\columnwidth]{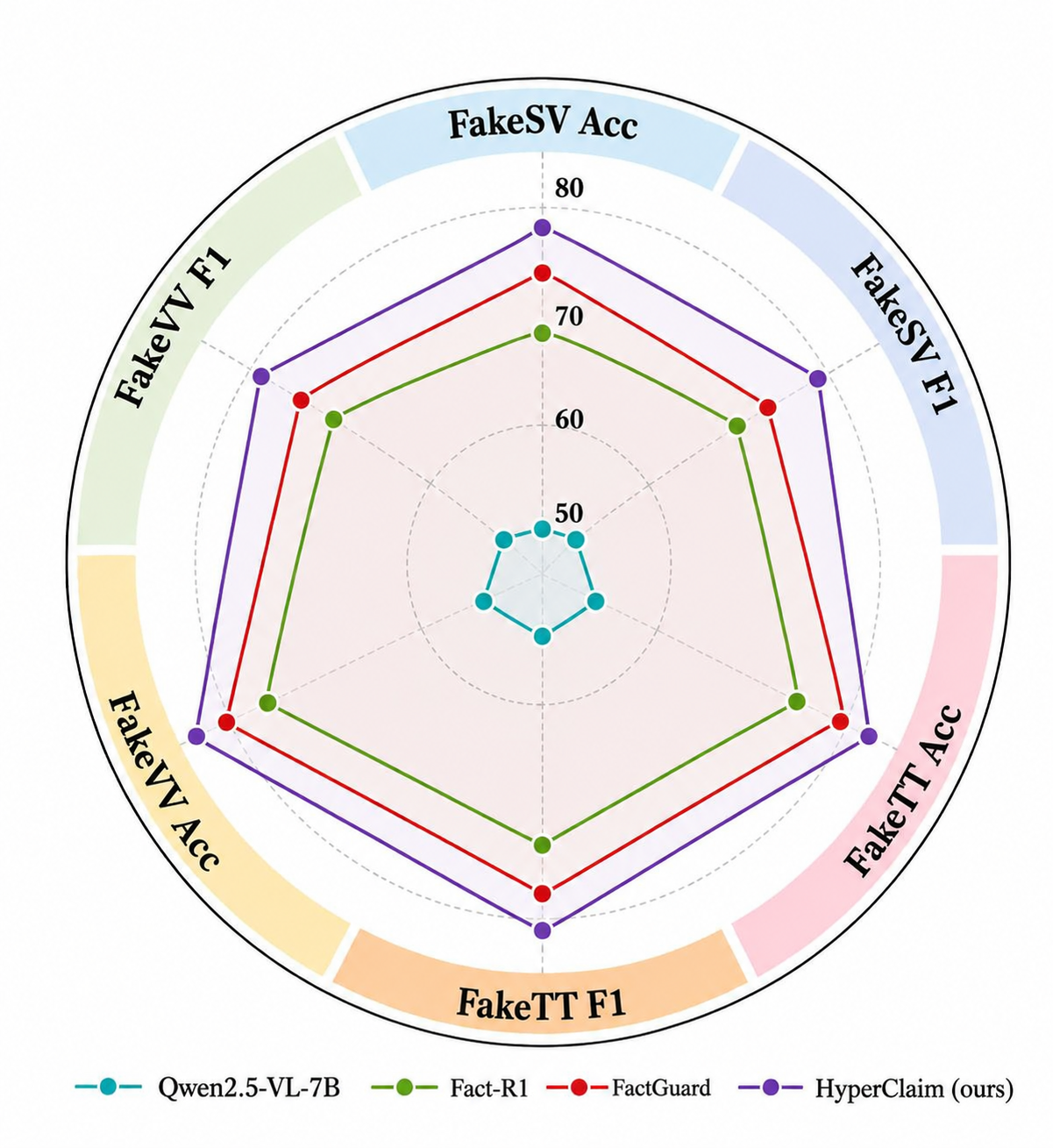}
    \caption{\method (ours) and representative baselines on FakeSV, FakeTT, and FakeVV.}
    \label{fig:radar}
\end{figure}

Existing video misinformation detectors exploit multimodal correlation, social context, creative-process cues, consistency modeling, and domain generalization~\cite{shang2021tiktok,qi2023fakesv,qi2023need,bu2024fakingrecipe,zong2024opinion,guo2025doctor,wang2025fakesvvlm}. Most condense heterogeneous evidence into modality- or sample-level representations. Although effective for capturing global semantics, this aggregation can dilute inconsistencies involving only a specific token, frame, or short temporal relation. Reasoning-centric systems pursue a different direction: Fact-R1 learns long-form multimodal reasoning through instruction tuning, preference optimization, and reinforcement learning~\cite{zhang2025factr1}, while FactGuard performs ambiguity-aware verification with selective tool use~\cite{li2026factguard}. These approaches are valuable for open-ended inference and externally grounded verification. We study a complementary closed-input setting in which the relevant signal is already present but distributed across the title, contextual text, and temporally ordered frames. The central challenge is to represent and aggregate this localized evidence structure without losing it through global fusion.

We propose \method, a fine-grained cross-modal hypergraph framework for video misinformation detection. The title or benchmark-provided paired text serves as a claim-like query that organizes the remaining evidence, while the prediction target remains the original real/fake label. Hypergraphs naturally represent higher-order evidence units involving a textual anchor and multiple related frames while preserving claim context, evidence-source structure, and short-range temporal relations~\cite{feng2019hgnn,bai2021hypergraph}. \hforge constructs sparse claim, evidence, temporal, and cross-modal hyperedges through confidence-aware filtering and claim-aware source budgeting. \aether calibrates text--video interactions, learns soft node--hyperedge memberships, and propagates information within and across evidence units. \cred aggregates textual, visual, and hyperedge states and models claim--video agreement and discrepancy for authenticity prediction. Closest to our work, HGTMFC performs fine-grained static image--text verification through hypergraph and line-graph propagation~\cite{pang2025beyondtext}; \method extends this principle to temporally ordered video with sparse query--text--frame construction and adaptive evidence routing. Learned incidence and attention weights further support token- and frame-level structural evidence tracing rather than free-form or causal explanation.

Our contributions are as follows:
\begin{itemize}
    \item We introduce a claim-oriented sparse temporal hypergraph representation that preserves localized query--text--frame and short-range temporal relations instead of collapsing heterogeneous evidence into global modality-level summaries.
    \item We develop \method, integrating confidence-aware hypergraph formation, adaptive soft-incidence reasoning, residual text--video calibration, and discrepancy-aware node- and hyperedge-level readout for real/fake prediction.
    \item Under the FactGuard temporal protocol, \method improves accuracy and F1 on FakeSV, FakeTT, and FakeVV, while supporting token- and frame-level structural evidence tracing.
\end{itemize}

\section{Related Work}

\paragraph{Video misinformation detection.}
Existing systems exploit multimodal clues, social context, creation cues, consistency, and domain generalization~\cite{shang2021tiktok,choi2021fanvm,qi2023fakesv,qi2023need,bu2024fakingrecipe,zong2024opinion,wang2025cafvd,guo2025doctor,li2026mva,wang2025fakesvvlm}. These methods establish the value of heterogeneous evidence but generally summarize it at the modality or sample level. \method instead preserves localized token--frame and short-range temporal relations.

\paragraph{Cross-modal evidence grounding.}
Multimodal misinformation can arise when each modality appears plausible in isolation but the modalities are jointly inconsistent. NewsCLIPpings and VERITE investigate out-of-context mismatches and unimodal shortcuts~\cite{luo2021newsclippings,papadopoulos2024verite}, while MOCHEG and subsequent surveys emphasize the retrieval and aggregation of claim-relevant multimodal evidence~\cite{yao2023mocheg,akhtar2023multimodal}. These studies show that effective verification requires more than global feature fusion, particularly when the decisive inconsistency is confined to a specific entity, event, or temporal cue. \method extends this perspective to video by preserving localized query--text--frame relations as explicit higher-order evidence units.

\paragraph{Hypergraph reasoning for multimodal verification.}
Hypergraph neural networks provide a general mechanism for representing relations involving more than two elements~\cite{feng2019hgnn,bai2021hypergraph}. Most closely related to our work, HGTMFC combines hypergraph and line-graph propagation to model fine-grained claim--evidence interactions in static image--text verification~\cite{pang2025beyondtext}. \method extends this line of research to temporally ordered video by introducing sparse claim-oriented construction, overlapping temporal hyperedges, adaptive soft incidence, residual text--video calibration, and discrepancy-aware readout.

\paragraph{Reasoning-centric verification.}
Fact-R1 learns long-form multimodal reasoning, whereas FactGuard combines ambiguity estimation, selective tool use, and reinforcement learning~\cite{zhang2025factr1,li2026factguard}. These approaches are valuable for open-ended or externally grounded verification. \method addresses the complementary closed-input setting by reasoning discriminatively over explicit local evidence structure.

\begin{figure*}[!t]
    \centering

    \includegraphics[width=\textwidth]{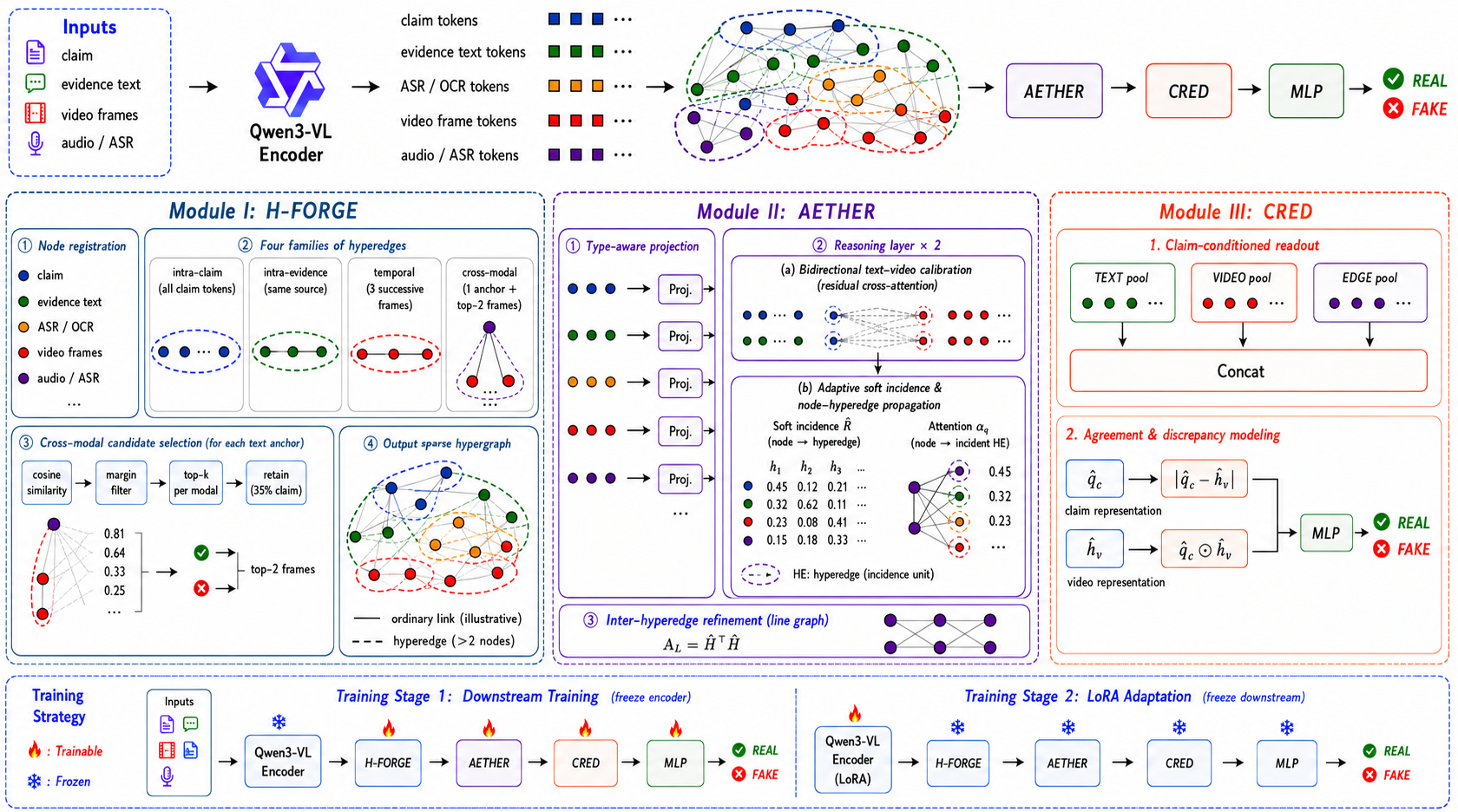}
    \caption{Overview of \method. \hforge constructs a sparse temporal heterogeneous hypergraph from claim-like query tokens, textual-evidence tokens, and sampled frames. Each \aether layer first performs lightweight bidirectional text--video calibration and then adaptive soft-incidence node--hyperedge reasoning; a line-graph layer subsequently refines interactions among evidence units. \cred pools textual, visual, and hyperedge states and predicts the original real/fake label from claim--video agreement and discrepancy. The bottom panel shows the two-stage optimization strategy.}
    \label{fig:pipeline}
\end{figure*}

\section{Methodology}

\subsection{Problem Formulation and Overview}

The evaluated benchmarks formulate video misinformation detection as sample-level authenticity classification rather than independently annotated claim--evidence entailment. Given a video-news sample $x=(q,E,v)$, where $q$ denotes its title or benchmark-provided paired text, $E$ contains the remaining textual context, and $v$ is the associated video, the goal is to predict the original benchmark label
\begin{equation}
 y\in\{\realnews,\fakenews\}.
\label{eq:task_label}
\end{equation}

We use the term \emph{claim} operationally to refer to the title-derived or benchmark-provided query $q$. It is the proposition around which the remaining multimodal content is organized, but it is not an independently annotated fact-checking claim. For FakeSV~\cite{qi2023fakesv} and FakeTT~\cite{bu2024fakingrecipe}, the video title is used as $q$, while available non-title fields---including comments, automatic speech recognition (ASR) transcripts, optical character recognition (OCR) text, keywords, and captions---form $E$. Each textual item is retained as a separate evidence source. For FakeVV~\cite{zhang2025factr1}, the benchmark-provided text paired with the video serves as $q$. Query and evidence tokens are assigned to disjoint node sets, and the query is not duplicated in $E$.

Accordingly, \method predicts the original real/fake label rather than a separate entailment label. Claim conditioning is used to prioritize and summarize title-relevant evidence; it does not assume that every fake sample is reducible to one explicit contradiction between the query and the observed content.

As illustrated in Figure~\ref{fig:pipeline}, \method contains four stages: (1) claim and evidence encoding; (2) sparse temporal hypergraph formation with \hforge; (3) adaptive temporal hypergraph reasoning with \aether; and (4) claim-conditioned authenticity prediction with \cred. The model is trained discriminatively using the original benchmark labels.

\paragraph{Design rationale.}
Dense text--frame fusion propagates redundancy, so \hforge retains a bounded set of reliable textual anchors and constructs explicit local evidence units. Similarity-based candidates need not be factually reliable, so \aether recalibrates cross-modal states and learns context-dependent soft memberships and message weights. Finally, \cred preserves complementary textual, visual, and edge-level summaries while exposing claim--video agreement and discrepancy. This decomposition separates candidate construction, evidence routing, and prediction instead of leaving all three to a single dense fusion block.

\subsection{Claim and Evidence Encoding}

We use Qwen3-VL-Embedding-2B~\cite{li2026qwen3vlembedding} as a unified multimodal encoder. For the claim-like query and each textual evidence item, we retain token-level hidden states in $\mathbb{R}^{d}$, where $d=2048$. Token-level representations are important because factual conflicts often depend on local entities, actions, numbers, and relations that may disappear under sentence-level pooling.

For the visual branch, we retain one frame every 30 original frames until either 16 frames have been collected or the video ends. The retained frames remain in chronological order, and each is represented by a frame-level visual embedding. This asymmetric representation keeps language evidence fine-grained while bounding the number of visual nodes. Let $\mathbf{C}=\{\mathbf{c}_k\}$, $\mathbf{X}=\{\mathbf{x}_i\}$, and $\mathbf{V}=\{\mathbf{v}_j\}_{j=1}^{T}$ denote the claim-token, evidence-token, and sampled-frame representations, respectively.

\subsection{\hforge: Claim-Oriented Sparse Hypergraph Formation}

We construct a heterogeneous hypergraph $\mathcal{G}=(\mathcal{V},\mathcal{H})$ with
\begin{equation}
\mathcal{V}=\mathcal{V}_c\cup\mathcal{V}_e\cup\mathcal{V}_v,
\label{eq:node_set}
\end{equation}
where $\mathcal{V}_c$, $\mathcal{V}_e$, and $\mathcal{V}_v$ contain claim tokens, evidence tokens, and sampled frames. Four complementary hyperedge families are instantiated:
\begin{equation}
\mathcal{H}=\mathcal{H}_c\cup\mathcal{H}_e\cup\mathcal{H}_v\cup\mathcal{H}_{tv}.
\label{eq:edge_set}
\end{equation}
$\mathcal{H}_c$ contains one intra-claim hyperedge connecting all tokens of $q$. Each item in $E$ contributes one hyperedge in $\mathcal{H}_e$ connecting all tokens encoded from that item; thus, individual comments, ASR segments, OCR entries, keywords, or captions remain source-specific. $\mathcal{H}_v$ preserves local temporal context with overlapping windows of three consecutive sampled frames and stride one. If fewer than three frames are available, all available frame nodes form one temporal hyperedge; boundary padding is not used. Finally, $\mathcal{H}_{tv}$ contains sparse cross-modal verification units connecting a textual anchor to its strongest frame matches.

\paragraph{Confidence-aware cross-modal filtering.}
A dense text--frame graph introduces many irrelevant links because short videos contain redundant frames and social text contains noisy evidence. \hforge therefore constructs $\mathcal{H}_{tv}$ through a claim-oriented filtering process. Let $\mathbf{u}_i$ denote a textual source node from $\mathcal{V}_c\cup\mathcal{V}_e$. For frame node $\mathbf{v}_j$, we compute
\begin{equation}
 s_{ij}=\frac{\mathbf{u}_i^{\top}\mathbf{v}_j}
 {\lVert\mathbf{u}_i\rVert_2\lVert\mathbf{v}_j\rVert_2}.
\label{eq:sim}
\end{equation}
Let $j_1(i)$ and $j_2(i)$ denote the highest- and second-highest-scoring frames. Define the best-match score $s_i^{\star}=s_{i,j_1(i)}$ and margin $m_i=s_{i,j_1(i)}-s_{i,j_2(i)}$. We retain
\begin{equation}
\mathcal S=\{i:s_i^{\star}\geq\delta_s,\;m_i\geq\delta_m\},
\label{eq:select_set}
\end{equation}
where $\delta_s=0.14$ and $\delta_m=0.003$. Retained anchors are ranked by
\begin{equation}
r_i=s_i^{\star}+\eta m_i,\qquad \eta=0.1.
\label{eq:rank_score}
\end{equation}
The first term measures cross-modal matching strength, while the margin term favors anchors whose best frame match is less ambiguous. This heuristic is fixed across all three datasets.

\paragraph{Claim-aware source budgeting.}
We use a textual-anchor budget of $B_s=48$ and reserve approximately 35\% of the slots (17 by default) for claim-token anchors; the remaining slots are assigned to evidence-token anchors. Claim priority is therefore introduced through source-aware budgeting rather than an additional ranking term. Within each source type, anchors are selected by $r_i$, and each retained anchor is connected to its two highest-scoring frames. The complete query is not copied into every cross-modal hyperedge: query context enters directly through claim-token anchors and indirectly through $\mathcal{H}_c$, \aether propagation, and the \cred readout.

\begin{figure*}[!t]
    \centering
    \includegraphics[width=\textwidth]{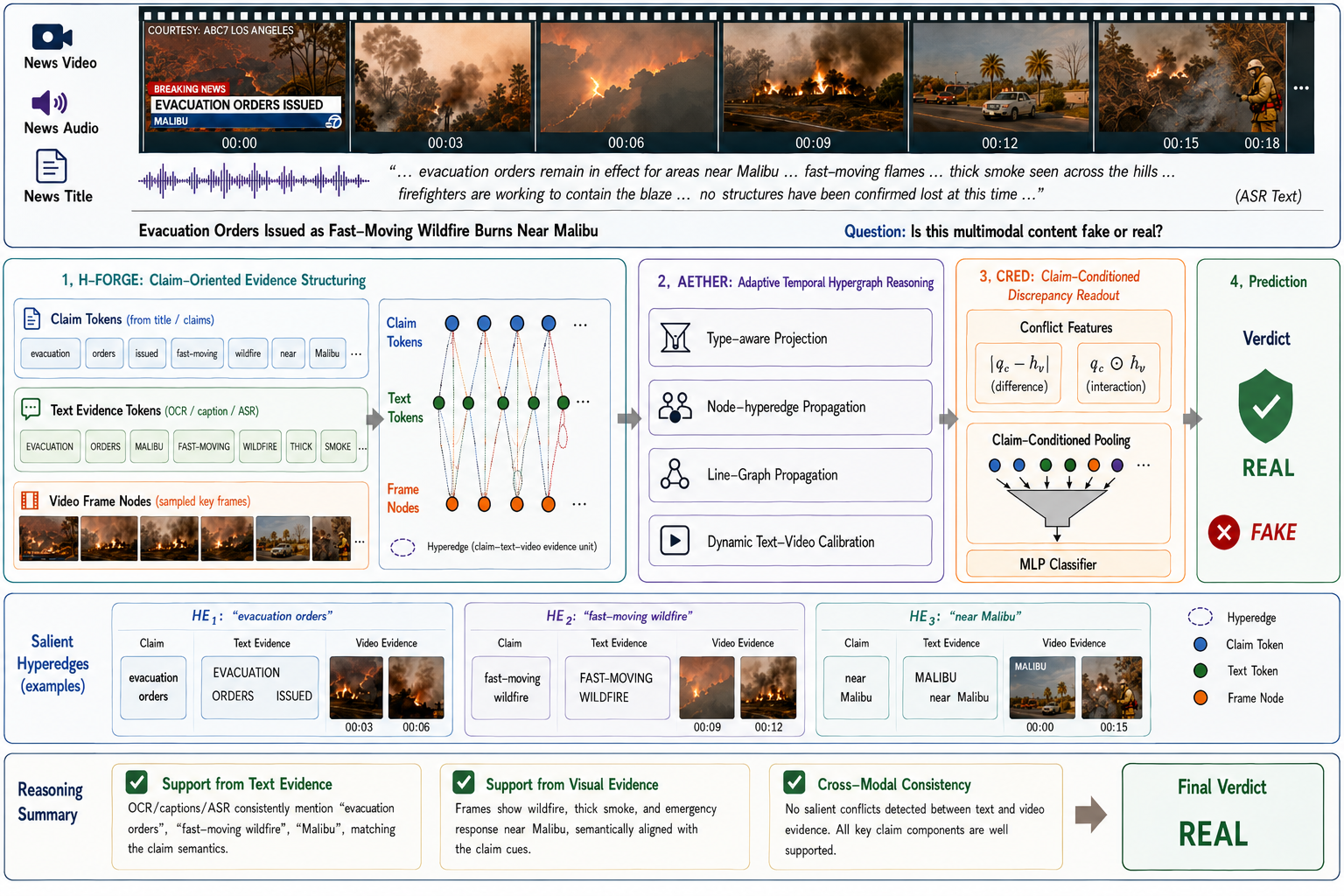}
    \caption{Structural evidence trace produced by \method. High-weight evidence units connect localized query phrases, textual cues, and temporally relevant frames.}
    \label{fig:evidence_trace}
\end{figure*}

\subsection{\aether: Adaptive Temporal Hypergraph Reasoning}

The initial graph contains plausible evidence units, but encoder similarity alone should not determine their final contribution. Node features are first projected by type-specific linear layers into a shared hidden space of dimension 512. \aether contains two reasoning layers, each combining lightweight bidirectional text--video calibration with adaptive node--hyperedge propagation; a final line-graph layer refines interactions among hyperedges.

\paragraph{Bidirectional text--video calibration.}
At reasoning layer $\ell$, a lightweight cross-attention block~\cite{vaswani2017attention} recalibrates the current textual and visual states:
\begin{equation}
\widetilde{\mathbf T}^{(\ell)},\widetilde{\mathbf V}^{(\ell)}
=\operatorname{XAttnBlock}
\big(\mathbf T^{(\ell)},\mathbf V^{(\ell)}\big).
\label{eq:tv_calibration}
\end{equation}
The block forms one scaled text--video affinity matrix. It is normalized over video nodes for messages to text and over text nodes for the reverse direction. Each attention distribution is mixed with a 0.05 uniform residual; the two messages are added with scale 0.35 and followed by Layer Normalization~\cite{ba2016layernorm}. The module uses one full-dimensional attention head, temperature 1.0, and no feed-forward sublayer. The calibrated states $\widetilde{\mathbf T}^{(\ell)}$ and $\widetilde{\mathbf V}^{(\ell)}$ replace the current states in the subsequent soft-incidence and node--hyperedge updates of the same layer. After the second layer, \cred pools the final calibrated textual and visual states. Thus, the calibration outputs are part of the prediction path rather than auxiliary representations, while the discrete candidate structure remains unchanged.

\paragraph{Soft incidence and node--hyperedge propagation.}
For hyperedge $h$ with members $\mathcal{N}(h)$, the seed state $\bar{\mathbf{h}}_h$ is the mean of its current member states. Let $\mathbf{q}_h=\mathbf{W}_q\bar{\mathbf{h}}_h$ and $\mathbf{k}_i=\mathbf{W}_k\mathbf{h}_i$. We compute
\begin{equation}
\begin{aligned}
e_{ih} &= \mathbf{q}_h^{\top}\mathbf{k}_i/\tau,\\
\alpha_{ih} &= \operatorname*{softmax}_{i\in\mathcal{N}(h)}(e_{ih}),\\
\widetilde{h}_{ih} &= (1-\rho)\alpha_{ih}
+\rho/|\mathcal{N}(h)|.
\end{aligned}
\label{eq:soft_incidence}
\end{equation}
Here $\tau=1.0$ and $\rho=0.10$. The matrix $\widetilde{\mathbf H}\in\mathbb{R}^{|\mathcal V|\times|\mathcal H|}$ reweights only the sparse candidate memberships constructed by \hforge.

A grouped aggregation module first forms member-group means and then applies attention to obtain the hyperedge state $\mathbf z_h$. Let $\mathbf q_i^n=\mathbf W_Q\mathbf h_i$ and $\mathbf k_h^e=\mathbf W_K\mathbf z_h$. Each node attends to its incident hyperedges using
\begin{equation}
a_{ih}=\operatorname*{softmax}_{h\in\mathcal N(i)}\!\left(\frac{(\mathbf q_i^n)^\top\mathbf k_h^e}{\sqrt{d_h}}+\log(\widetilde h_{ih}+\epsilon)\right).
\label{eq:node_edge_attn}
\end{equation}
The implementation uses a hidden size of 512 and eight attention heads. The resulting message is combined with the node state using residual weight $\beta=0.5$, Layer Normalization, and dropout 0.1. Soft incidence is recomputed in each reasoning layer.

\paragraph{Inter-hyperedge refinement.}
After the two node--hyperedge layers, we construct a weighted line graph
\begin{equation}
\mathbf{A}_L=\widetilde{\mathbf{H}}^{\top}\widetilde{\mathbf{H}}.
\label{eq:line_graph}
\end{equation}
We remove its diagonal and apply one degree-normalized GCN-style layer~\cite{kipf2017gcn} with a linear projection, ReLU, residual connection, and Layer Normalization. This allows overlapping evidence units to exchange information without collapsing their individual states.

\begin{table*}[!t]
\centering
\setlength{\tabcolsep}{1mm}
\begin{tabular}{lrrrrrrrrrrrr}
\toprule
\multirow{2}{*}{Model} & \multicolumn{4}{c}{FakeSV} & \multicolumn{4}{c}{FakeTT} & \multicolumn{4}{c}{FakeVV} \\
\cmidrule(lr){2-5}\cmidrule(lr){6-9}\cmidrule(lr){10-13}
& Acc & Prec & Rec & F1 & Acc & Prec & Rec & F1 & Acc & Prec & Rec & F1 \\
\midrule
BERT$^{\dagger}$ & 65.4 & 66.0 & 66.5 & 66.2 & 68.7 & 67.5 & 67.5 & 67.5 & 60.4 & 57.9 & 56.8 & 57.3 \\
ViT$^{\dagger}$ & 57.5 & 57.9 & 58.1 & 58.0 & 59.3 & 59.1 & 59.5 & 59.3 & 55.5 & 59.3 & 58.7 & 59.0 \\
TikTec$^{\dagger}$ & 64.8 & 63.2 & 61.9 & 62.5 & 61.1 & 64.8 & 64.2 & 64.5 & 59.3 & 59.1 & 59.5 & 59.3 \\
FANVM$^{\dagger}$ & 65.4 & 66.1 & 64.3 & 65.2 & 68.9 & 64.7 & 68.8 & 67.1 & 61.9 & 60.7 & 60.8 & 60.8 \\
SV-FEND$^{\dagger}$ & 67.1 & 67.4 & 66.3 & 66.8 & 67.6 & 72.2 & 69.0 & 70.6 & 70.9 & 71.4 & 71.3 & 71.3 \\
FakingRec$^{\dagger}$ & 69.5 & 69.7 & 70.4 & 70.0 & 71.0 & 71.9 & 72.0 & 72.0 & 72.1 & 72.4 & 71.6 & 72.0 \\
\midrule
Gemini2-thinking$^{\dagger}$ & 63.1 & 61.8 & 61.9 & 61.9 & 56.6 & 55.2 & 55.3 & 55.3 & 51.5 & 46.0 & 46.0 & 48.6 \\
GPT-4o$^{\dagger}$ & 66.6 & 65.2 & 64.7 & 64.9 & 57.9 & 57.8 & 62.9 & 63.7 & 56.0 & 60.4 & 35.0 & 44.3 \\
GPT-o1-mini$^{\dagger}$ & 60.3 & 57.7 & 56.5 & 57.1 & 52.5 & 51.6 & 51.7 & 51.7 & 47.5 & 46.9 & 37.6 & 41.8 \\
DeepSeek-R1$^{\dagger}$ & 61.8 & 60.4 & 60.3 & 60.3 & 49.8 & 52.6 & 52.5 & 52.6 & 53.5 & 58.1 & 25.2 & 35.1 \\
Qwen2.5-VL-7B$^{\dagger}$ & 55.6 & 55.5 & 55.7 & 55.6 & 54.9 & 54.0 & 54.1 & 54.0 & 52.9 & 51.1 & 51.1 & 51.1 \\
Qwen2.5-VL-72B$^{\dagger}$ & 57.6 & 55.4 & 55.2 & 55.3 & 59.2 & 58.1 & 58.3 & 58.2 & 54.0 & 60.0 & 24.0 & 34.3 \\
QVQ-72B-preview$^{\dagger}$ & 60.8 & 59.0 & 58.8 & 58.9 & 58.1 & 54.0 & 52.8 & 53.4 & 53.5 & 52.6 & 52.6 & 52.6 \\
InternVL2.5-8B$^{\dagger}$ & 49.8 & 52.6 & 52.5 & 52.6 & 43.9 & 44.0 & 44.0 & 44.0 & 53.5 & 58.5 & 24.0 & 34.0 \\
InternVL2.5-78B-MPO$^{\dagger}$ & 57.5 & 53.0 & 52.0 & 52.5 & 59.2 & 57.1 & 56.7 & 56.9 & 54.0 & 60.0 & 24.0 & 34.3 \\
\midrule
Fact-R1$^{\dagger}$ & 75.6 & 77.7 & 72.0 & 74.7 & 74.4 & 77.8 & 68.3 & 72.7 & 81.2 & 84.5 & 76.4 & 80.3 \\
FactGuard$^{\ddagger}$ & 79.3 & 82.2 & 80.6 & 81.4 & 75.3 & 73.8 & 76.7 & 75.2 & 83.0 & 85.8 & 82.1 & 83.9 \\
\textbf{\method (ours)} & \textbf{83.7} & \textbf{84.6} & \textbf{85.3} & \textbf{84.2} & \textbf{82.0} & \textbf{81.3} & \textbf{77.9} & \textbf{79.5} & \textbf{87.3} & \textbf{88.7} & \textbf{84.8} & \textbf{86.1} \\
\bottomrule
\end{tabular}
\caption{Performance (\%) on FakeSV, FakeTT, and FakeVV.}
\label{tab:main}
\parbox{0.99\textwidth}{\small $^{\dagger}$From Fact-R1~\cite{zhang2025factr1}; $^{\ddagger}$from FactGuard~\cite{li2026factguard}; baseline results were not rerun.}
\end{table*}

\subsection{\cred: Claim-Conditioned Discrepancy Readout}

\cred operates on the final node states and line-graph-refined hyperedge states. The claim representation is the mean of the final claim-token states:
\begin{equation}
\mathbf{q}_c=\operatorname{MeanPool}
\left(\{\mathbf{h}_i:i\in\mathcal{V}_c\}\right).
\label{eq:claim_query}
\end{equation}
Textual nodes and all hyperedges of the sample are summarized by learned linear attention, whereas visual nodes are pooled by attention conditioned on $\mathbf q_c$. For each source $m\in\{t,v,e\}$, the corresponding summary has the common form
\begin{equation}
\mathbf{h}_m=\sum_{i}\pi_i^{(m)}\mathbf{s}_i^{(m)},
\label{eq:claim_conditioned_readout}
\end{equation}
where $\mathbf{s}_i^{(t)}=\mathbf h_i$, $\mathbf{s}_i^{(v)}=\mathbf h_i$, and $\mathbf{s}_i^{(e)}=\mathbf z_i$. Text and hyperedge scores are produced by learned linear scorers; visual scores use the dot product between projected $\mathbf q_c$ and projected frame states. All pooling temperatures are 1.0. Let $\mathbf d_{cv}=|\widehat{\mathbf q}_c-\widehat{\mathbf h}_v|$ and $\mathbf p_{cv}=\widehat{\mathbf q}_c\odot\widehat{\mathbf h}_v$, where the hats denote $\ell_2$ normalization. The classifier input is
\begin{equation}
\mathbf z=[\mathbf h_t;\mathbf h_v;\mathbf h_e;\mathbf d_{cv};\mathbf p_{cv}]\in\mathbb R^{2560}.
\label{eq:conflict_feature}
\end{equation}
The difference term exposes claim--video discrepancy, whereas the element-wise product represents feature-level agreement. The verdict distribution is produced by a multilayer perceptron (MLP):
\begin{equation}
p(y\mid q,E,v)=\operatorname{MLP}(\mathbf{z}),
\label{eq:classifier}
\end{equation}
where the classifier comprises
$\operatorname{Linear}(2560,512)$--GELU~\cite{hendrycks2016gelu}--
Dropout$(0.1)$--$\operatorname{Linear}(512,2)$.

\subsection{Structural Evidence Tracing}

\method does not generate free-form rationales. Instead, learned incidence $\widetilde{h}_{ih}$, node--hyperedge attention $a_{ih}$, and readout weights expose high-weight information paths that can be mapped back to query tokens, evidence tokens, and frame timestamps. We visualize these paths as structural evidence traces. They indicate routing salience rather than causal or faithfulness-certified explanations.

\subsection{Training Objective}

The model is optimized with cross-entropy:
\begin{equation}
\mathcal{L}_{\mathrm{cls}}=-\log p(y\mid q,E,v).
\label{eq:loss}
\end{equation}
We adopt a two-stage optimization strategy. In Stage~1, Qwen3-VL-Embedding-2B~\cite{li2026qwen3vlembedding} is frozen, and the hypergraph reasoning, pooling, and prediction modules are optimized on precomputed multimodal features. In Stage~2, all downstream modules---including \hforge, \aether, \cred, the pooling scorers, and the classifier---remain frozen. Only LoRA adapters~\cite{hu2022lora} in both the visual and textual attention and feed-forward projections of the multimodal encoder are optimized on raw multimodal inputs. Stage~1 establishes a stable evidence-routing and decision function, while Stage~2 adapts encoder representations to this fixed downstream structure without perturbing the learned reasoning path.

\section{Experiments}
\subsection{Experimental Settings}

\paragraph{Datasets and evaluation.}
We evaluate on FakeSV~\cite{qi2023fakesv}, FakeTT~\cite{bu2024fakingrecipe}, and FakeVV~\cite{zhang2025factr1}. FakeSV/FakeTT use titles as queries and non-title fields as evidence; FakeVV uses its paired text as the query. Following FactGuard~\cite{li2026factguard}, the chronologically latest 15\% of each dataset form the test set. We report accuracy, precision, recall, and F1 following Fact-R1~\cite{zhang2025factr1}.

\paragraph{Baselines.}
We compare \method with conventional discriminative systems, general-purpose multimodal/reasoning models, and task-aligned reasoning systems. The conventional group contains BERT, ViT, TikTec, FANVM, SV-FEND, and FakingRec~\cite{devlin2019bert,dosovitskiy2021vit,shang2021tiktok,choi2021fanvm,qi2023fakesv,bu2024fakingrecipe}. The general-purpose group contains Gemini2-thinking, GPT-4o, GPT-o1-mini, DeepSeek-R1, Qwen2.5-VL-7B/72B, QVQ-72B-preview, and InternVL2.5-8B/78B-MPO\@. Their values, together with Fact-R1, are taken from the unified comparison in Fact-R1~\cite{zhang2025factr1}; FactGuard values are taken from its corresponding evaluation~\cite{li2026factguard}. For models without native video input, Fact-R1 uses news-domain video descriptions as textual surrogates. The comparison therefore covers text-only, vision-only, multimodal fusion, zero-shot multimodal reasoning, and task-specific verification paradigms, while the shared temporal split controls the principal evaluation protocol.

\paragraph{Implementation details.}
We use Qwen3-VL-Embedding-2B~\cite{li2026qwen3vlembedding}, sample one frame every 30 raw frames (maximum 16), and construct three-frame windows with stride one. \hforge uses $(\delta_s,\delta_m,\eta)=(0.14,0.003,0.1)$, at most 48 anchors, a 35\% claim quota, and two frame matches per anchor. \aether uses two 512-dimensional eight-head node--hyperedge layers and one line-graph layer, with $(\tau_h,\rho_h,\beta)=(1.0,0.10,0.5)$ and $(\tau_{tv},\rho_{tv},\lambda_{tv})=(1.0,0.05,0.35)$. Readout temperature is 1.0; \cred uses a 512-dimensional GELU classifier with dropout 0.1. AdamW uses weight decay $10^{-4}$~\cite{loshchilov2019adamw}. Stage~1 learning rates are $10^{-4}$ for hypergraph modules and $5\times10^{-4}$ for pooling/classification, with effective batch size 32. Stage~2 trains only Qwen LoRA parameters at $2\times10^{-5}$ with effective batch size 16 on one A100. Both stages use three epochs of linear warmup, cosine annealing to $10^{-7}$~\cite{loshchilov2017sgdr}, and gradient clipping at 1.0. LoRA~\cite{hu2022lora} is applied to visual/textual attention and feed-forward projections with $r=\alpha=32$ and dropout 0.05.

\subsection{Main Results}

\paragraph{Overall performance.}
As summarized in Figure~\ref{fig:radar} and detailed in Table~\ref{tab:main}, \method achieves the highest accuracy and F1 on all three benchmarks. It reaches 83.7\% accuracy and 84.2\% F1 on FakeSV, 82.0\% accuracy and 79.5\% F1 on FakeTT, and 87.3\% accuracy and 86.1\% F1 on FakeVV\@. The consistency of these gains is important because the benchmarks differ in language, platform, scale, and construction procedure, suggesting that localized high-order evidence modeling is not restricted to a single entity-replacement dataset.

\paragraph{Comparison with task-aligned reasoning systems.}
Compared with Fact-R1, \method improves accuracy by 8.1, 7.6, and 6.1 percentage points on FakeSV, FakeTT, and FakeVV, respectively. Relative to FactGuard, the gains are 4.4, 6.7, and 4.3 points. Notably, the largest margin over FactGuard occurs on FakeTT, showing that the improvement is not confined to FakeVV's entity-replacement construction. These results do not imply that discriminative reasoning universally dominates open-world agentic verification. Rather, under the shared closed-benchmark setting, explicit local evidence structure provides a strong decision path without making the final verdict contingent on generated reasoning trajectories or tool calls.

\paragraph{General models and dataset effects.}
Zero-shot multimodal large language models remain substantially weaker than task-aligned systems; for example, GPT-4o obtains 56.0\% accuracy and 44.3\% F1 on FakeVV\@. The strongest \method result also occurs on FakeVV, whose entity-replacement construction aligns with the localized relations targeted by \hforge~\cite{zhang2025factr1}. Improvements on FakeSV and FakeTT indicate that the benefit extends beyond this construction.

\subsection{Ablation Studies}

\paragraph{Input evidence.}
Table~\ref{tab:abl_modality} evaluates evidence composition on FakeSV\@. Text alone provides a relatively strong signal, but adding video increases F1 from 70.1\% to 80.5\%. ASR evidence alone contributes less than visual evidence, whereas the full configuration is strongest. This pattern supports the central motivation: the decision cannot be reduced to textual plausibility, and the full model benefits from jointly modeling modalities rather than merely concatenating them.

\begin{table}[t]
\centering
\small
\setlength{\tabcolsep}{4pt}
\begin{tabular*}{\columnwidth}{@{\extracolsep{\fill}}lcccc@{}}
\toprule
Variant & \ablmetric{Acc} & \ablmetric{Prec} & \ablmetric{Rec} & \ablmetric{F1} \\
\midrule
Text only & 70.4 & 70.8 & 69.6 & 70.1 \\
Text + Video & 80.5 & 80.9 & 79.4 & 80.5 \\
Text + ASR & 73.2 & 73.6 & 72.1 & 72.7 \\
\textbf{Full model} (\textit{w/o LoRA}) & \textbf{81.6} & \textbf{82.1} & \textbf{80.3} & \textbf{81.2} \\
\bottomrule
\end{tabular*}
\caption{Multi-modal Combination Ablation; Values are percentages.}
\label{tab:abl_modality}
\end{table}

\paragraph{Claim-conditioned readout and discrepancy modeling.}
Table~\ref{tab:abl_readout} isolates the prediction stage on FakeSV\@. Replacing learned aggregation with global mean pooling reduces F1 to 73.2\%, indicating that evidence selection must preserve source- and claim-dependent relevance. Claim-conditioned visual attention improves the result even without discrepancy features. Hard top-$k$ pooling is stronger than mean pooling but remains below the full model, showing that soft evidence weighting is preferable to committing to a fixed subset. The complete \cred module reaches 81.2\% F1 by combining learned textual/edge pooling, claim-conditioned visual pooling, and explicit claim--video agreement and discrepancy.
\begin{table}[t]
\centering
\small
\setlength{\tabcolsep}{3.3pt}

\begin{tabular*}{\columnwidth}
{@{\extracolsep{\fill}}lccc|cccc@{}}
\toprule
 & \multicolumn{3}{c|}{Component}
 & \multicolumn{4}{c}{Metrics (\%)} \\
\cmidrule(lr){2-4}\cmidrule(l){5-8}
Variant & CQ & DF & AP & Acc & Prec & Rec & F1 \\
\midrule
Mean readout
& \xmark & \xmark & \xmark
& 74.3 & 74.8 & 73.0 & 73.2 \\

Without discrepancy
& \cmark & \xmark & \cmark
& 77.1 & 78.7 & 76.5 & 77.4 \\

Hard top-$k$ pooling
& \cmark & \cmark & \xmark
& 79.4 & 79.9 & 78.6 & 78.9 \\

\textbf{Full \cred} (\textit{w/o LoRA})
& \cmark & \cmark & \cmark
& \textbf{81.6} & \textbf{82.1}
& \textbf{80.3} & \textbf{81.2} \\
\bottomrule
\end{tabular*}

\caption{CQ: claim query; DF: discrepancy features;
AP: attention pooling.}
\label{tab:abl_readout}
\end{table}
\paragraph{Adaptive evidence routing.}
Table~\ref{tab:abl_graph} evaluates the two adaptive components of \aether on FakeSV\@. A static candidate hypergraph reaches 75.7\% F1. Learning soft incidence improves F1 to 78.0\% by suppressing unreliable candidate memberships, while residual text--video calibration reaches 79.1\% by recovering complementary continuous alignment. Combining both mechanisms yields 81.2\% F1, indicating that sparse structured propagation and continuous calibration are complementary rather than interchangeable.

\begin{table}[t]
\centering
\small
\setlength{\tabcolsep}{3.3pt}

\begin{tabular*}{\columnwidth}{@{\extracolsep{\fill}}lcc|cccc@{}}
\toprule
 & \multicolumn{2}{c|}{Component}
 & \multicolumn{4}{c}{Metrics (\%)} \\
\cmidrule(lr){2-3}\cmidrule(l){4-7}
Variant & SI & TVC & Acc & Prec & Rec & F1 \\
\midrule
Static hypergraph
& \xmark & \xmark & 76.1 & 76.6 & 74.8 & 75.7 \\
Soft incidence only
& \cmark & \xmark & 78.5 & 79.0 & 77.1 & 78.0 \\
Calibration only
& \xmark & \cmark & 79.6 & 80.1 & 78.2 & 79.1 \\
\textbf{Full \aether} (\textit{w/o LoRA})
& \cmark & \cmark
& \textbf{81.6} & \textbf{82.1}
& \textbf{80.3} & \textbf{81.2} \\
\bottomrule
\end{tabular*}

\caption{SI: soft incidence; TVC: text--video calibration.}
\label{tab:abl_graph}
\end{table}

\paragraph{Hyperparameter ablations.}
Detailed hyperparameter ablations of H-Forge, Aether, adaptive mechanisms, and input regularization are provided in the supplementary material.

\paragraph{Qualitative structural evidence tracing.}
Figure~\ref{fig:evidence_trace} visualizes high-weight hyperedges and
routing paths that connect query phrases such as ``evacuation
orders,'' ``fast-moving wildfire,'' and ``near Malibu'' with relevant
transcript/OCR cues and video frames. Compared with a single
video-level importance score, these traces localize the textual and
visual elements emphasized during prediction. They reflect learned
routing salience and should not be interpreted as causal or
faithfulness-certified explanations.

\section{Conclusion}

We introduced \method, a sparse temporal hypergraph framework for
video misinformation detection. It preserves localized query--text--frame
and temporal relations through claim-oriented construction, adaptive
node--hyperedge reasoning, and discrepancy-aware readout. Across three
benchmarks, \method improves accuracy and F1 while supporting structural
evidence tracing.

\appendix
\section{Supplementary Material}
\label{sec:supplement}

This supplement reports implementation details, reproducibility information, and controlled hyperparameter analyses for \method. Unless stated otherwise, all ablations use FakeSV \citep{qi2023fakesv} without low-rank adaptation (LoRA) \citep{hu2022lora} and report accuracy (Acc), precision (Prec), recall (Rec), and F1-score (F1) as percentages. The sweeps characterize local sensitivity around the canonical operating point rather than global optimality. Motivated by prior short-video work on multimodal evidence and cross-modal consistency \citep{qi2023fakesv,qi2023need,wang2025fakesvvlm}, we examine evidence sparsity, propagation depth, and learned cross-modal adaptation.

\section{Implementation and Reproducibility Details}
\label{sec:supp_impl}

\paragraph{Feature extraction.}
We use Qwen3-VL-Embedding-2B \citep{li2026qwen3vlembedding} as a unified multimodal encoder. For the claim-like query and each textual evidence item, we retain token-level hidden states in $\mathbb{R}^{d_{\mathrm{enc}}}$, where $d_{\mathrm{enc}}=2048$. For video, we sample one frame every 30 raw frames and retain at most 16 frames per instance. Frame-level representations are extracted with the same encoder, ensuring that textual and visual inputs are represented within a common semantic space. Type-specific linear projections then map query-token, evidence-token, and frame features into a shared 512-dimensional hidden space. The projections are not shared across node types, which allows the model to preserve modality-specific statistics while enforcing a common dimensionality for subsequent message passing.

\paragraph{Candidate hypergraph construction.}
Hypergraphs provide a natural representation for multiway relations that cannot be reduced to a single pairwise edge without loss of structure; this motivation underlies a range of hypergraph neural architectures \citep{feng2019hgnn,yadati2019hypergcn,chien2022allset}. In \method, the heterogeneous candidate hypergraph $\mathcal{G}=(\mathcal{V},\mathcal{E})$ contains three node types: query (claim) tokens $\mathcal{V}_q$, evidence tokens $\mathcal{V}_e$, and sampled frames $\mathcal{V}_f$. We instantiate four hyperedge families: (i) claim-to-anchor hyperedges that connect the query to selected textual anchors; (ii) intra-textual hyperedges that preserve local evidence structure; (iii) cross-modal hyperedges that associate textual anchors with matched video frames; and (iv) temporal hyperedges that connect neighboring frames using a window of size 3 and stride 1.

The default \hforge configuration retains at most $B_s=48$ anchors, allocates 35\% of the anchor budget to claim-token anchors, and attaches the top-2 frame matches to each textual anchor. Cross-modal filtering uses $(\delta_s,\delta_m,\eta)=(0.14,0.003,0.1)$. These values define a sparse candidate structure; they do not by themselves determine the final contribution of each node--hyperedge relation, which is subsequently refined by the adaptive incidence mechanism.

\paragraph{Hypergraph reasoning.}
\aether contains two 512-dimensional, eight-head node--hyperedge propagation layers followed by one line-graph layer. The node--hyperedge updates alternate aggregation from nodes to hyperedges and from hyperedges back to nodes, consistent with the general message-passing view adopted by modern hypergraph networks \citep{feng2019hgnn,chien2022allset}. A residual coefficient $\beta=0.5$ combines hyperedge-derived messages with the current node states. The soft-incidence module uses $(\tau_h,\rho_h)=(1.0,0.10)$, while text--video calibration uses $(\tau_{tv},\rho_{tv},\lambda_{tv})=(1.0,0.05,0.35)$. Learning or refining incidence structure is motivated by the fact that a constructed hypergraph can contain noisy, task-irrelevant, or missing relations \citep{cai2022hsl}; in our model, the candidate structure serves as an inductive prior rather than an immutable graph.

\paragraph{Readout and classifier.}
\cred performs claim-conditioned attention over the calibrated textual, visual, and hyperedge states. The readout temperature is $\tau_r=1.0$. The aggregated representation is passed to a 512-dimensional classifier with a Gaussian Error Linear Unit (GELU) nonlinearity \citep{hendrycks2016gelu} and dropout probability 0.1 \citep{srivastava2014dropout}.

\paragraph{Optimization.}
We train \method with AdamW \citep{loshchilov2019adamw} and weight decay $10^{-4}$. For the ablations reported here, the learning rate is $10^{-4}$ for the hypergraph modules and $5\times10^{-4}$ for pooling and classification, with an effective batch size of 32. Training begins with three epochs of linear warmup and then follows cosine annealing to a terminal learning rate of $10^{-7}$, using the standard cosine-scheduling family \citep{loshchilov2017sgdr}. Gradients are clipped to a maximum norm of 1.0.

\paragraph{Execution environment and reporting protocol.}
The implementation is based on PyTorch \citep{paszke2019pytorch} and is executed on a single NVIDIA A100 GPU\@. All rows in the reported sweeps use the same fixed random seed and the same default settings for every non-ablated factor. Consequently, the tables provide controlled point estimates under a common protocol; they do not provide run-to-run uncertainty or statistical significance estimates.

\section{Ablation Protocol and Interpretation}
\label{sec:supp_ablations}

We vary one hyperparameter per subtable while holding all others fixed. This protocol isolates local sensitivity but neither estimates factorial interactions nor establishes global optima. Because every row is a fixed-seed point estimate, numerical differences are interpreted descriptively, and mechanistic explanations are presented only as hypotheses consistent with the trends. We emphasize F1 while retaining all metrics; boldface marks the highest F1 in each block and $\dagger$ the canonical default. Across the sweeps, the central trade-off is between evidence coverage and structural selectivity.

\subsection{H-Forge: Constructing Sparse Evidence Units}
\label{sec:supp_hforge}

\hforge determines which textual and visual elements enter the candidate hypergraph. Because the downstream reasoning layers operate on this structure, changes in anchor coverage or cross-modal connectivity affect both the available evidence and the routes through which messages can propagate. The goal of this stage is therefore not maximal connectivity, but selective coverage of claim-relevant evidence.

\subsubsection{Anchor budget and claim-aware evidence quota}

The anchor budget $B_s$ limits the number of textual spans that can seed evidence units. Anchors serve as cross-modal hubs, so increasing $B_s$ simultaneously expands textual coverage and the number of potential anchor--frame connections.

Table~\ref{tab:supp_hforge}(a) shows a non-monotonic response. Increasing the budget from $B_s=24$ to $B_s=48$ improves F1 from 78.8\% to 81.2\% ($+2.4$ points) and recall from 76.2\% to 80.3\% ($+4.1$ points). This pattern is consistent with under-coverage at the smallest budget: the candidate graph may not retain enough secondary evidence to support claim verification. Increasing the budget further to $B_s=64$ reduces F1 to 80.5\% and precision from 82.1\% to 80.3\%. The latter change is consistent with a reduction in structural precision when lower-ranked anchors are admitted, although the table alone does not directly measure anchor quality.

At fixed $B_s=48$, the 35\% claim-aware quota also occupies an interior optimum. Reducing the quota to 20\% lowers F1 by 1.4 points and recall by 3.1 points relative to the default, whereas increasing it to 50\% lowers F1 by 1.1 points. Thus, within the tested range, retaining a moderate amount of non-anchor textual context is preferable to either aggressive pruning or broad evidence retention. Together, the two controls implement complementary forms of sparsification: $B_s$ limits the number of evidence hubs, while the quota limits the contextual breadth around those hubs.

\subsubsection{Cross-modal filtering thresholds}

After textual evidence units are formed, \hforge applies semantic and mismatch criteria to candidate anchor--frame links. The pair $(\delta_s,\delta_m)$ therefore controls which visual nodes are eligible to participate in cross-modal hyperedges, while $\eta$ is fixed at 0.1 throughout this sweep.

Table~\ref{tab:supp_hforge}(b) again exhibits a non-monotonic pattern. At $\delta_m=0.003$, the default $\delta_s=0.14$ reaches 81.2\% F1, exceeding $\delta_s=0.10$ and $\delta_s=0.18$ by 1.5 and 1.3 points, respectively. At $\delta_s=0.14$, the default $\delta_m=0.003$ exceeds $\delta_m=0.001$ by 1.7 F1 points and $\delta_m=0.005$ by 0.8 points. Notably, the $(0.14,0.005)$ setting obtains slightly higher precision (82.2\%) but lower recall (79.9\%) and F1 (80.4\%) than the default. This metric divergence indicates that the thresholds affect not only the amount of retained visual evidence but also the balance between conservative and inclusive predictions.

The selected pair $(0.14,0.003)$ is therefore the strongest point in the evaluated grid under F1. More generally, the results support a selective-alignment regime in which neither endpoint of the tested threshold range is preferred. We deliberately limit this conclusion to the evaluated grid rather than treating the selected pair as a globally optimal threshold combination.

\begin{table}[t]
\centering
\caption{\hforge hyperparameter sensitivity on FakeSV without LoRA.}
\label{tab:supp_hforge}
\small
\begin{tabular}{lcccc}
\toprule
Configuration & Acc & Prec & Rec & F1 \\
\midrule
\multicolumn{5}{l}{\textbf{(a) Anchor budget $B_s$ and claim-aware quota}} \\
$B_s=24$, quota 35\% & 79.4 & 79.7 & 76.2 & 78.8 \\
$B_s=32$, quota 35\% & 80.7 & 81.0 & 77.6 & 80.2 \\
$B_s=48$, quota 35\%\defaultmark & 81.6 & 82.1 & 80.3 & \textbf{81.2} \\
$B_s=64$, quota 35\% & 81.1 & 80.3 & 79.1 & 80.5 \\
$B_s=48$, quota 20\% & 80.6 & 80.5 & 77.2 & 79.8 \\
$B_s=48$, quota 50\% & 80.8 & 81.1 & 78.4 & 80.1 \\
\midrule
\multicolumn{5}{l}{\textbf{(b) Cross-modal thresholds $(\delta_s,\delta_m)$}} \\
(0.10, 0.003) & 80.4 & 79.4 & 77.1 & 79.7 \\
(0.14, 0.001) & 80.8 & 80.9 & 77.4 & 79.5 \\
(0.14, 0.003)\defaultmark & 81.6 & 82.1 & 80.3 & \textbf{81.2} \\
(0.14, 0.005) & 81.0 & 82.2 & 79.9 & 80.4 \\
(0.18, 0.003) & 80.5 & 80.4 & 77.5 & 79.9 \\
\bottomrule
\end{tabular}
\end{table}

\subsection{Aether: Reasoning Over the Heterogeneous Hypergraph}
\label{sec:supp_aether}

\aether propagates information over the candidate structure produced by \hforge. We study three aspects of its effective receptive field: propagation depth, the number of visual matches attached to each textual anchor, and the temporal extent of frame-level hyperedges.

\subsubsection{Number of reasoning layers}

Table~\ref{tab:supp_aether_struct}(a) varies the number of node--hyperedge reasoning layers. A single layer obtains 79.1\% F1, whereas two layers improve F1 to 81.2\% ($+2.1$ points). The improvement supports the need for more than one propagation step when information must pass among claim tokens, evidence anchors, and visual nodes.

A third layer slightly increases accuracy from 81.6\% to 81.7\% and precision from 82.1\% to 82.9\%, but decreases recall from 80.3\% to 79.6\% and F1 from 81.2\% to 80.8\%. Thus, two layers provide the strongest precision--recall balance even though they do not maximize every individual metric. The decline at greater depth is consistent with the broader observation that repeated graph propagation can smooth node representations and attenuate local distinctions \citep{li2018deeper}; however, the present ablation does not directly measure representation similarity, so over-smoothing should be understood as a plausible explanation rather than a demonstrated mechanism.

\subsubsection{Frame matches per textual anchor}

Table~\ref{tab:supp_aether_struct}(b) controls the cross-modal neighborhood size of each textual anchor. Expanding from the top-1 to the top-2 matched frames increases F1 from 79.7\% to 81.2\% and recall from 77.2\% to 80.3\%. This indicates that a single visual match does not provide sufficient coverage for the best-performing configuration.

The gains saturate after two matches. Top-3 and top-4 matching reduce F1 to 80.7\% and 80.4\%, respectively. Since the added frames are lower-ranked under the matching function, this trend is consistent with diminishing evidence quality as the neighborhood expands. The empirical conclusion is limited but clear: among the tested fixed neighborhood sizes, top-2 matching provides the best balance between visual coverage and cross-modal selectivity.

\subsubsection{Temporal hyperedge window size}

Table~\ref{tab:supp_aether_struct}(c) varies the local temporal window used to construct frame--frame hyperedges. A window of 2 yields 80.0\% F1. Expanding the window to 3 improves F1 by 1.2 points, while a further expansion to 5 reduces F1 to 80.6\%. The intermediate window therefore provides the strongest result within the tested range.

This pattern supports local rather than global temporal aggregation. A very narrow window limits the temporal context available to each matched frame, whereas a broader window increases the chance that distinct visual moments are pooled into the same higher-order relation. Although the ablation does not identify the specific events responsible for the change, it establishes that the temporal receptive field is a material design choice rather than an inconsequential implementation detail.

\begin{table}[t]
\centering
\caption{\aether depth and structural sensitivity on FakeSV without LoRA.}
\label{tab:supp_aether_struct}
\small
\begin{tabular}{lcccc}
\toprule
Configuration & Acc & Prec & Rec & F1 \\
\midrule
\multicolumn{5}{l}{\textbf{(a) Number of \aether reasoning layers}} \\
1 layer & 79.7 & 80.2 & 76.5 & 79.1 \\
2 layers\defaultmark & 81.6 & 82.1 & 80.3 & \textbf{81.2} \\
3 layers & 81.7 & 82.9 & 79.6 & 80.8 \\
\midrule
\multicolumn{5}{l}{\textbf{(b) Frame matches per textual anchor}} \\
Top-1 frame & 80.4 & 79.4 & 77.2 & 79.7 \\
Top-2 frames\defaultmark & 81.6 & 82.1 & 80.3 & \textbf{81.2} \\
Top-3 frames & 81.4 & 80.5 & 79.4 & 80.7 \\
Top-4 frames & 81.1 & 80.2 & 79.0 & 80.4 \\
\midrule
\multicolumn{5}{l}{\textbf{(c) Temporal hyperedge window size}} \\
Window size 2 & 80.7 & 78.8 & 77.5 & 80.0 \\
Window size 3\defaultmark & 81.6 & 82.1 & 80.3 & \textbf{81.2} \\
Window size 5 & 81.2 & 81.8 & 79.8 & 80.6 \\
\bottomrule
\end{tabular}
\end{table}

\subsection{Adaptive Structure and Cross-Modal Calibration}
\label{sec:supp_adaptive}

\aether combines two forms of adaptation. Soft incidence modifies the strength of node--hyperedge membership relative to the candidate structure, while text--video calibration adjusts continuous node representations using cross-modal agreement. This separation mirrors a useful distinction in adaptive hypergraph learning: the model can refine \emph{which relations are trusted} and \emph{how the participating features are calibrated} \citep{cai2022hsl}.

\subsubsection{Soft-incidence parameters}

The temperature $\tau_h$ controls the scale of the learned membership scores, and the residual gate $\rho_h$ controls the contribution of the learned update relative to the candidate incidence. Table~\ref{tab:supp_adaptive}(a) shows that $(\tau_h,\rho_h)=(1.0,0.10)$ yields the highest F1 within the evaluated grid.

At fixed $\tau_h=1.0$, disabling the learned update with $\rho_h=0$ reduces F1 from 81.2\% to 79.7\% ($-1.5$ points), indicating that the static candidate incidence is not sufficient for the strongest result. A smaller nonzero gate ($\rho_h=0.05$) reaches 80.5\% F1, whereas a larger gate ($\rho_h=0.20$) reaches 80.6\%. These values show that refinement is beneficial, but the tested setting with the largest update is not the best one; the candidate structure remains a useful prior.

At fixed $\rho_h=0.10$, changing $\tau_h$ from 1.0 to 0.5 or 2.0 reduces F1 by 1.0 and 1.8 points, respectively. The central temperature is therefore preferred within this sweep. This conclusion is purely empirical and does not rely on assigning ``sharper'' or ``flatter'' behavior to a temperature value, which depends on the exact normalization convention used in the implementation.

\subsubsection{Text--video calibration parameters}

The calibration weight $\rho_{tv}$ controls the magnitude of the cross-modal correction, while $\lambda_{tv}$ controls residual mixing with the original state; $\tau_{tv}=1.0$ is held fixed in this sweep. Removing calibration by setting $\rho_{tv}=0$ lowers F1 to 79.5\%, 1.7 points below the canonical setting $(0.05,0.35)$. Similarly, reducing $\lambda_{tv}$ to 0.20 yields 79.9\% F1. These comparisons show that a nontrivial calibration pathway contributes to the reported performance.

Importantly, Table~\ref{tab:supp_adaptive}(b) does \emph{not} identify the canonical setting as the numerical F1 maximizer. Increasing $\lambda_{tv}$ to 0.50 yields 81.6\% F1, and increasing $\rho_{tv}$ to 0.10 yields 81.8\% F1. The latter is 0.6 points above the canonical setting. We retain $(0.05,0.35)$ as the default because it is the fixed configuration used to define the operating point for the remaining ablations and the main experimental pipeline, not because this single-run sweep establishes it as uniquely optimal. In the absence of repeated-run uncertainty estimates, the 0.4--0.6 point differences among these high-performing settings should be treated as descriptive rather than statistically resolved. Overall, the sweep indicates a reasonably broad effective region for calibration, with clear degradation only when the calibration path is removed or substantially weakened.

\begin{table}[t]
\centering
\caption{Adaptive-mechanism sensitivity on FakeSV without LoRA.}
\label{tab:supp_adaptive}
\small
\begin{tabular}{lcccc}
\toprule
Configuration & Acc & Prec & Rec & F1 \\
\midrule
\multicolumn{5}{l}{\textbf{(a) Soft-incidence parameters $(\tau_h,\rho_h)$}} \\
(0.5, 0.10) & 80.7 & 80.4 & 77.4 & 80.2 \\
(1.0, 0.00) & 80.9 & 80.5 & 77.1 & 79.7 \\
(1.0, 0.05) & 81.1 & 81.3 & 78.0 & 80.5 \\
(1.0, 0.10)\defaultmark & 81.6 & 82.1 & 80.3 & \textbf{81.2} \\
(1.0, 0.20) & 81.2 & 81.8 & 79.1 & 80.6 \\
(2.0, 0.10) & 80.8 & 80.9 & 77.5 & 79.4 \\
\midrule
\multicolumn{5}{l}{\textbf{(b) Calibration parameters $(\rho_{tv},\lambda_{tv})$}} \\
(0.00, 0.35) & 80.7 & 80.2 & 77.3 & 79.5 \\
(0.05, 0.20) & 80.5 & 80.0 & 77.1 & 79.9 \\
(0.05, 0.35)\defaultmark & 81.6 & 82.1 & 80.3 & 81.2 \\
(0.05, 0.50) & 81.2 & 80.9 & 79.8 & 81.6 \\
(0.10, 0.35) & 81.4 & 81.7 & 79.4 & \textbf{81.8} \\
\bottomrule
\end{tabular}
\end{table}

\subsection{Input Granularity and Regularization}
\label{sec:supp_input_reg}

The final group of ablations examines how much visual information enters the model and how strongly the classifier is regularized. These settings affect computational redundancy, evidence coverage, and the concentration of the final claim-conditioned readout.

\subsubsection{Frame sampling density}

Table~\ref{tab:supp_input_reg}(a) compares temporal stride and frame-cap settings. With a maximum of 16 frames, sampling every 30 raw frames reaches 81.2\% F1. Denser sampling every 15 frames yields 80.8\% F1, while sparser sampling every 60 frames yields 79.7\% F1. The comparatively small 0.4-point difference between strides 15 and 30 suggests that the denser sequence is largely redundant under the fixed 16-frame cap, whereas the 1.5-point reduction at stride 60 indicates a greater risk of omitting informative moments.

At a fixed stride of 30, reducing the frame cap from 16 to 8 lowers F1 by 1.2 points. Increasing the cap to 24 also lowers F1, but only by 0.5 points. These results identify the 16-frame cap as the strongest tested compromise between coverage and redundancy. Because video duration and frame rate can vary across instances, the conclusion should be interpreted as specific to the adopted preprocessing pipeline rather than as a universal sampling rate.

\subsubsection{Dropout rate}

Table~\ref{tab:supp_input_reg}(b) evaluates dropout in the final classifier. Removing dropout reduces F1 from 81.2\% to 80.3\%. Increasing the rate to 0.2 and 0.3 yields 80.7\% and 80.1\% F1, respectively. Thus, light regularization is beneficial, but stronger dropout progressively reduces the retained task signal. This behavior is consistent with the standard role of dropout as a regularizer that must be balanced against representational capacity \citep{srivastava2014dropout}.

\subsubsection{Readout temperature}

Table~\ref{tab:supp_input_reg}(c) varies the temperature $\tau_r$ used by the claim-conditioned attention readout in \cred. The default $\tau_r=1.0$ reaches 81.2\% F1, compared with 80.0\% at $\tau_r=0.5$ and 80.5\% at $\tau_r=2.0$. The intermediate setting therefore provides the best tested balance. As with the soft-incidence temperature, the empirical result is independent of whether a larger numerical value corresponds to a sharper or flatter distribution under the implementation's exact scaling convention.

\begin{table}[t]
\centering
\caption{Input and regularization sensitivity on FakeSV without LoRA.}
\label{tab:supp_input_reg}
\small
\begin{tabular}{lcccc}
\toprule
Configuration & Acc & Prec & Rec & F1 \\
\midrule
\multicolumn{5}{l}{\textbf{(a) Frame sampling density}} \\
Every 15 frames, max 16 & 81.4 & 81.9 & 79.4 & 80.8 \\
Every 30 frames, max 16\defaultmark & 81.6 & 82.1 & 80.3 & \textbf{81.2} \\
Every 60 frames, max 16 & 80.3 & 80.4 & 77.1 & 79.7 \\
Every 30 frames, max 8 & 80.7 & 80.5 & 77.5 & 80.0 \\
Every 30 frames, max 24 & 81.3 & 81.5 & 79.5 & 80.7 \\
\midrule
\multicolumn{5}{l}{\textbf{(b) Dropout rate}} \\
0.0 & 80.9 & 81.4 & 77.6 & 80.3 \\
0.1\defaultmark & 81.6 & 82.1 & 80.3 & \textbf{81.2} \\
0.2 & 81.3 & 81.6 & 79.2 & 80.7 \\
0.3 & 80.8 & 80.4 & 78.5 & 80.1 \\
\midrule
\multicolumn{5}{l}{\textbf{(c) Readout temperature $\tau_r$}} \\
0.5 & 80.7 & 80.1 & 77.4 & 80.0 \\
1.0\defaultmark & 81.6 & 82.1 & 80.3 & \textbf{81.2} \\
2.0 & 81.1 & 81.5 & 79.0 & 80.5 \\
\bottomrule
\end{tabular}
\end{table}

\section{Cross-Ablation Synthesis}
\label{sec:supp_interactions}

The one-factor-at-a-time design does not estimate statistical interactions between hyperparameters. Nevertheless, the marginal trends are mutually consistent and identify a common operating regime.

\paragraph{Moderate evidence density and shallow propagation.}
The anchor sweep favors $B_s=48$ over both smaller and larger budgets, while the depth sweep favors two layers over one or three. Considered together, these independent results support a hypergraph that is sufficiently connected for multi-step evidence integration but not so dense or deep that weak relations are repeatedly propagated. This is a cross-ablation synthesis, not a direct test of the $B_s\times$depth interaction.

\paragraph{Selective cross-modal matching with local temporal support.}
The strongest frame-matching setting attaches two frames to each textual anchor, and the strongest temporal setting uses a window of three frames. Both sweeps therefore favor small local neighborhoods over either minimal or broader connectivity. The resulting design separates two roles: anchor--frame links select visually relevant moments, whereas temporal hyperedges supply limited context around those moments.

\paragraph{Candidate structure and learned adaptation.}
Soft incidence and text--video calibration operate at different representational levels. The former adjusts node--hyperedge membership, while the latter modifies continuous cross-modal states. As reported in Table~\ref{tab:abl_graph}, removing either mechanism decreases F1 relative to the full configuration (from 81.2\% to 79.1\% without soft incidence and to 78.0\% without calibration). These component ablations indicate that neither mechanism is redundant. They do not, by themselves, establish a formal interaction effect or identify a unique causal decomposition of the gain.

\section{Consolidated Default Configuration}
\label{sec:supp_default}

For reproducibility, Table~\ref{tab:supp_defaults} consolidates the settings that define the canonical ablation operating point. The table distinguishes between the \emph{canonical default} and the numerically best point in every isolated sweep; as discussed above, these coincide for most factors but not for the calibration sweep.

\begin{table}[t]
\centering
\caption{Canonical configuration used for the controlled ablations.}
\label{tab:supp_defaults}
\footnotesize
\begin{tabular}{@{}p{0.29\columnwidth}p{0.64\columnwidth}@{}}
\toprule
Component & Setting \\
\midrule
Encoder & Qwen3-VL-Embedding-2B; encoder dimension 2048; projected hidden dimension 512 \\
Frame input & One frame per 30 raw frames; maximum 16 frames \\
\hforge sparsity & $B_s=48$ anchors; 35\% claim-aware evidence quota \\
Cross-modal links & Top-2 frame matches; $(\delta_s,\delta_m,\eta)=(0.14,0.003,0.1)$ \\
Temporal links & Window size 3; stride 1 \\
\aether backbone & Two 512-dimensional, eight-head node--hyperedge layers; one line-graph layer; $\beta=0.5$ \\
Soft incidence & $(\tau_h,\rho_h)=(1.0,0.10)$ \\
Calibration & $(\tau_{tv},\rho_{tv},\lambda_{tv})=(1.0,0.05,0.35)$ \\
\cred readout & $\tau_r=1.0$; 512-dimensional GELU classifier; dropout 0.1 \\
Optimization & AdamW; weight decay $10^{-4}$; learning rates $10^{-4}$ and $5\times10^{-4}$; effective batch size 32 \\
Schedule/stability & Three warmup epochs; cosine annealing to $10^{-7}$; gradient clipping at 1.0 \\
Hardware & Single NVIDIA A100 GPU \\
\bottomrule
\end{tabular}
\end{table}

\section{Summary}
\label{sec:supp_summary}

The supplementary analyses support a consistent empirical principle: \method performs best in a sparse, locally connected regime that preserves claim-relevant evidence without indiscriminately expanding the reasoning graph. In \hforge, this regime is represented by $B_s=48$, a 35\% claim-aware quota, and the tested cross-modal threshold pair $(0.14,0.003)$. In \aether, it is represented by two reasoning layers, top-2 anchor--frame matching, and temporal windows of size 3. Soft-incidence learning improves over a static candidate structure, and nonzero text--video calibration improves over removing the calibration path. The input analyses further favor moderate frame sampling, light dropout, and an intermediate readout temperature.

These conclusions are intentionally local to FakeSV, the no-LoRA setting, the evaluated parameter grids, and the fixed-seed protocol. Most canonical defaults maximize F1 within their respective sweeps. The calibration default is the principal exception: nearby settings yield modestly higher point estimates, indicating a high-performing region rather than a uniquely identified optimum. Taken together, the results substantiate the architectural emphasis on fine-grained, claim-conditioned cross-modal reasoning while avoiding claims of global optimality or statistical significance that are not supported by the present experiments.

\bibliography{references}

\end{document}